\documentclass{article}

\usepackage{times}
\usepackage{graphicx} 
\usepackage{subcaption}
\usepackage{cleveref}
\usepackage{booktabs}

\usepackage{tikz}
\usetikzlibrary{shapes,arrows,shapes.geometric}

\usepackage{natbib}

\usepackage{algorithm} \usepackage{algorithmic}

\usepackage{hyperref}

\usepackage{verbatim}
\usepackage{multirow}
\usepackage{makecell}

\usepackage[percent]{overpic}

\usepackage[accepted]{icml2017_arxiv_port} 

\usepackage[draft]{todonotes}

\icmltitlerunning{A Framework For Testing Intuitive Physics}

\begin{document}

\twocolumn[ \icmltitle{Probing Physics Knowledge Using Tools from Developmental Psychology}



	\icmlsetsymbol{equal}{*}

	\begin{icmlauthorlist} \icmlauthor{Luis Piloto}{dm,pr,equal} \icmlauthor{Ari
		Weinstein}{dm,equal} \icmlauthor{Dhruva T.B.}{dm} 
		  \icmlauthor{Arun Ahuja}{dm}  \icmlauthor{Mehdi Mirza}{dm} \icmlauthor{Greg Wayne}{dm}  \icmlauthor{David Amos}{dm} 
		\icmlauthor{Chia-chun Hung}{dm} 
		\icmlauthor{Matthew Botvinick}{dm}
    \end{icmlauthorlist}	
		

	\icmlaffiliation{dm}{DeepMind, London, UK} \icmlaffiliation{pr}{Princeton
	University, New Jersey, USA}

	\icmlcorrespondingauthor{Luis Piloto}{piloto@google.com}

	\icmlkeywords{physics, deep learning, variational autoencoder, KL-divergence, intuitive physics}

	\vskip 0.3in ]



\printAffiliationsAndNotice{} 

\begin{abstract} 
	In order to build agents with a rich understanding of their environment, one key objective is to endow them with a grasp of intuitive physics; an ability to reason about three-dimensional objects, their dynamic interactions, and responses to forces. While some work on this problem has taken the approach of building in components such as ready-made physics engines, other research aims to extract general physical concepts directly from sensory data. In the latter case, one challenge that arises is evaluating the learning system. Research on intuitive physics knowledge in children has long employed a violation of expectations (VOE) method to assess children's mastery of specific physical concepts. We take the novel step of applying this method to artificial learning systems. In addition to introducing the VOE technique, we describe a set of probe datasets inspired by classic test stimuli from developmental psychology. We test a baseline deep learning system on this battery, as well as on a physics learning dataset (``IntPhys'') recently posed by another research group. Our results show how the VOE technique may provide a useful tool for tracking physics knowledge in future research.

\end{abstract}   

\section{Introduction}
\label{introduction}
The goal of developing artificial agents that display an intuitive grasp of everyday physics -- a basic understanding of the behavior of objects and forces -- has been widely acknowledged as a core challenge in artificial intelligence (AI)~\cite{Denil2017, Ullman2017}. 
In many settings, the aim is to deploy artificial agents in physical environments, whether real or simulated. In such contexts, physics understanding is of obvious importance, not only because it supports effective interaction with objects, but also because it supports sample-efficient learning and broad transfer~\cite{Wu2015, Finn2017}. For related reasons, research in cognitive science has flagged intuitive physics as a ‘core domain’ of conceptual knowledge~\cite{Spelke2007}, emerging early in life~\cite{Spelke1994} and providing the foundation for concepts in many other domains~\cite{Bremner2015, Hespos2012}. 

Research on intuitive physics has been divided between two strategies. In some work, the approach has been to directly leverage physics knowledge, for example by integrating a ready-made physics engine into the agent architecture~\cite{Wu2017}. In other work, the goal has instead been to learn physics from raw sensor data~\cite{Denil2017, Botvinick2017}. The latter case -- our focus in the present work -- carries a special challenge, which is to evaluate what knowledge the agent has acquired. When the goal is to learn physics, it is essential to have some method of tracking the course of learning and measuring its success. 

One widely applied method for assessing physics knowledge is to train systems to produce explicit predictions concerning the evolution of physical systems~\cite{Lerer2016, Battaglia2016, Chang2017}, and to use the accuracy of such predictions as a measure of physics knowledge. While this technique can be quite powerful, it must be applied with delicacy, since both successes and failures of prediction can be tricky to interpret. Inaccurate predictions can derive from quantitative (e.g., metrical) errors rather than flawed conceptual knowledge.  For instance, mean-squared error in pixel-space would penalize inaccuracies  in predictions for large objects more than small objects, but this would not reflect the importance of size in physical \emph{concepts}. Conversely, successful prediction generally relies jointly on a set of physical concepts, making it difficult to crisply analyze the content of acquired knowledge. 

In the present work we introduce and evaluate a complementary method for probing physics knowledge in artificial systems. Our approach is directly inspired by research in developmental psychology. In particular, we adopt two key ideas. The first is to assess physics knowledge by targeting specific physical concepts. Developmental psychology has, over the past fifty years, identified a core set of principles, which together provide a foundation for intuitive physics.  These include concepts such as `object persistence', the fact that objects (even when occluded) do not wink in or out of existence; 
`unchangeableness', that size, shape, pattern, color do not spontaneously change;
`continuity', that moving objects, unless perturbed, will follow smooth trajectories; 
`solidity', that solid objects cannot interpenetrate; and `containment', that objects remain in their containers when containers are moved~\cite{Baillargeon2012}.


In addition to isolating these basic principles, developmental psychology has also invented and refined what is by now a widely accepted and replicated experimental technique for probing their acquisition, referred to as the \textit{violation of expectations} (VOE) method~\cite{Baillargeon1985}.  Here, children are presented with dynamic physical displays, some of which violate a basic physical principle, essentially taking the form of `magic tricks' in which objects disappear, float unsupported, or otherwise contravene physical law. The critical measure is surprise, typically quantified in terms of looking time. When a child stares longer at a display that violates a physical principle than a carefully matched display that does not, this provides evidence that the child understands the relevant principle~\cite{Ball1973}. 

In the present work, we introduce the VOE method as a tool for assessing intuitive physics understanding in AI systems.  In particular, we make three contributions. First, we formalize the technique, leveraging psychological notions of surprise that afford a direct quantitative link with AI methods. Second, we introduce a collection of procedurally generated datasets, closely modeled on work in developmental psychology. Finally, we apply the VOE method to a natural baseline model, establishing a benchmark for next-step research. Our results indicate that systems with intuitive physics knowledge may not be far out of reach, but that important challenges nonetheless remain. 

\section{The Violation of Expectation Paradigm}  


As noted above, in developmental psychology the dependent measure usually employed in VOE studies is looking time, which is well established to correlate with subjective surprise. Itti and Baldi~(\citeyear{Baldi2010}) demonstrate a link between gaze time, and gaze direction to a specific computational measure of surprise. In particular, they showed that experimentally observed looking times are well predicted by the estimated Kullback-Leibler divergence (KL) between prior expectations and posterior beliefs given perceptual inputs. This was further reinforced in subsequent work by~\citet{Teglas2011}.

Our novel contribution is to capitalize on this preceding work by applying the same measure to AI systems.  Specifically, we look at the KL between the latent prior and posterior distributions in suitably designed systems, and treat this as developmental psychologists treat looking times. This approach requires that the AI system under consideration supports calculation of the KL. Fortunately, the literature provides a general class of models that have this property, based on deep variational inference (discussed further in section~\ref{model}).



\section{Dataset and Probe Design} \label{datasets}

In order to apply the VOE method to AI systems, we developed an initial collection of datasets, taking direct inspiration from experiments in developmental psychology. This corpus is described in general terms, and we then provide details of five specific datasets. The full corpus will be made available online (for sample videos, see \href{https://goo.gl/M1Heho}{https://goo.gl/M1Heho}). 

\subsection{General Dataset Design}
The datasets we introduce consist of videos that are procedurally generated and run in the Mujoco physics engine~\cite{Todorov2012}.  Videos cover approximately 3 seconds of simulated time and are created in categories focusing on specific intuitive physics concepts, with each category separated into testing, validation, and training  sets.  Test and validation corpora comprise paired \emph{consistent} (plausible) and \emph{inconsistent} (implausible) probes.  The training corpus for each category consists of \emph{examples} and \emph{controls}, each discussed below.

Paired probes form the basis of experimentation in the VOE paradigm (see Figure~\ref{fig:test_data}).  Both paired probes are identical aside from a manipulation in the inconsistent probe which makes the scene physically implausible, and is carefully controlled to isolate particular concepts in intuitive physics.  In order to ensure the system is reasoning over the history of the scene (and not, for example, simply encoding the statistics of single frames), we ensure that the manipulation is never immediately observable.  This is usually accomplished by having the manipulation occur while occluded. 

In addition to test probes, we produce a training corpus which is made of \emph{consistent examples} as well as \emph{controls}.  The consistent examples are similar to the consistent probes but have a much greater degree of variability (see Figure~\ref{fig:train_data}).  Controls are designed to mitigate potential biases in the dataset, as illustrated through specific examples below.   

Choices made in designing the train and test data were  guided by three desiderata: diversity, control/matching, and train-test separation.  Diversity includes all forms of variability that are introduced in the datasets.  These include altering object shape, size, pose, event timing, floor pattern, and any other aspect selected during procedural generation.  Often, variability introduced in consistent training examples is so large that constraints imposed on the probe data are overstepped (for example, an intended occluder may not always occlude a target object in train data).  Variability was maximized among probes as well, but subject to the constraint that the resulting video  must demonstrate the phenomenon of interest (or violate it, in cases of inconsistent probes) 

The addition of controls ensures there is training data that matches the inconsistent probes after the manipulation.  This, however, is done in a manner that is consistent with physics (so generally controls begin differently from inconsistent probes but end similarly).  Because of the inclusion of controls in the training data, the model should not be surprised by an inconsistent probe simply because the scene ends in a manner different from consistent examples, and instead be surprised by the manipulation due to the context established earlier in the video.

Additionally, we counter-balance probes to ensure that the model cannot be surprised by some superficial aspect of the probe data (for an example, see section~\ref{object_persistence}). This means each dataset contains two types of consistent and inconsistent probe distributions.

To summarize, our dataset is characterized by different distributions that produce train and consistent examples, as well as carefully considered controls that push the model to recognize the difference between scenes that can be superficially similar but semantically quite different. In combination they represent a concerted effort to introduce significant train/test separation and require rich generalization by the model.

We now describe key aspects of the specific datasets employed in our experiments.  

\subsection{`Object Persistence'}
\label{object_persistence}

Perhaps the most fundamental aspect of intuitive physics is  understanding that objects cannot disappear from existence, and is often called `object persistence' (or `permanence').  Taking inspiration from a classic behavioral experiment~\cite{Baillargeon1985}, probes for this category involve a rigid plank falling on an object.  In the consistent probe, when the plank falls on the object, the plank occludes it while also remaining propped up by it as expected (Figure~\ref{fig:test_data}$a$, top row).  By contrast, in the inconsistent probe the plank falls on top of the object (in a manner that is initially identical) but ends up flat on the floor, as if the object had disappeared (Figure~\ref{fig:test_data}$a$, bottom row).  In the counterbalanced probes, the consistent probe has the plank falling flat on the floor in an otherwise empty scene, and the inconsistent probe has the plank falling in the same empty scene but ends up inexplicably propped up by an item that is made to appear under the plank while it occludes part of the floor. 

Variability in the consistent examples arises from using an object which is either a sphere or cube, and of differing sizes. The width and height of the plank is allowed to vary greatly as well.  Furthermore, both items have randomized initial poses so that in each video the plank falls in a manner that is unique and difficult to predict (Figure~\ref{fig:train_data}$a$, top row).  

In the controls, we must ensure the system does not learn simply to associate the initial presence of a cube (or sphere) with the plank remaining supported, so controls consist of videos where the plank falls on an open location in the floor while the additional object is visible but does not support the plank (Figure~\ref{fig:train_data}$a$, bottom row).  Another control has the plank falling on the floor with no other object present, perceptually matching the inconsistent probes after the manipulation.

\subsection{`Unchangeableness'}

By the principle of `Unchangeableness'~\cite{Baillargeon2012}, objects tend to retain their features (e.g., color, shape) over time. In the consistent probes of this dataset, a random
assortment of static objects are aligned in the foreground (Figure~\ref{fig:test_data}$b$, top row).  A screen is lowered in front of those objects, and is then raised.  The training examples unfold in the same manner, but objects may be scattered anywhere in the scene, and the screen has more variable size and may appear anywhere from the foreground to the background (Figure~\ref{fig:train_data}$b$).  This means that only some (or none) of the objects may be occluded, and may be occluded at different times depending on the vertical position of the screen.  Controls resemble training examples with either objects or the screen omitted.

The concept of unchangeableness relates to a number of different aspects of objects, and therefore we include multiple inconsistent probes.  The first set of inconsistent probes examines the unchangeableness of position of static objects, and moves the position of one or more of the objects while occluded.  The other two inconsistent probes manipulate either the color or shape (see Figure~\ref{fig:test_data}$b$, bottom row) of one or more objects while occluded.

\subsection{`Continuity'}

The concept that an object traces out one continuous path through space and time is referred to in the developmental literature as `continuity'~\cite{Spelke1994}.  For videos in this category we use a nearly identical setup to a classic experiment~\cite{Aguilar1999} where consistent probes begin with two static pillars separated by a gap (Figure~\ref{fig:test_data}$c$, top row).  A ball is rolled horizontally behind both pillars so that it is visible before, between, and after the pillars during its trajectory.  In the inconsistent probes, while the ball is occluded by the first pillar, the ball is made invisible for the period when it would be between the pillars, and then reappears after the second pillar (Figure~\ref{fig:test_data}$c$, bottom row).

Consistent examples (Figure~\ref{fig:train_data}$c$, top row) are equivalent to consistent probes, but the ball is not constrained to be on a trajectory that is occluded by the pillars.  In one form of control (Figure~\ref{fig:train_data}$c$, bottom row) there is an additional large occluding block that fills the gap between the pillars, resulting in the ball being visible in only the same locations as in an inconsistent probe.

\subsection{`Solidity'}

This dataset is a recreation of an experiment that uses an object and an occluder~\cite{Hespos2001b} to test understanding of the solidity of objects, as related to the penetration of an object through a container and the ground below.  In probes, perspective is carefully controlled such that the camera can view inside the top of the container but not the bottom.  In consistent probes (Figure~\ref{fig:test_data}$d$, top row) a rectangular block is dropped into the container and comes to rest as expected.  In the inconsistent probes (Figure~\ref{fig:test_data}$d$, bottom row), the object ``falls through'' the container and the floor, and therefore disappears from view (with the penetration itself occluded by the face of the container) even if the object should remain visible due to its height.  Consistent examples (Figure~\ref{fig:train_data}$d$, top row) are similar to the probes but do not control for perspective; objects range from short enough to be completely occluded by the edge of the container to tall enough to protrude from the top of the container.  Controls include scenes where no object is visible, which perceptually matches the inconsistent probes after the object falls out of view, as well as scenes where the object falls behind the container (Figure~\ref{fig:train_data}$d$ bottom row, so that it is occluded by the container, but in a different manner).

\subsection{`Containment'}

\begin{figure}[t!]
    \centering
    \begin{subfigure}[]{\linewidth}
        \centering
        \begin{overpic}[width=\textwidth]{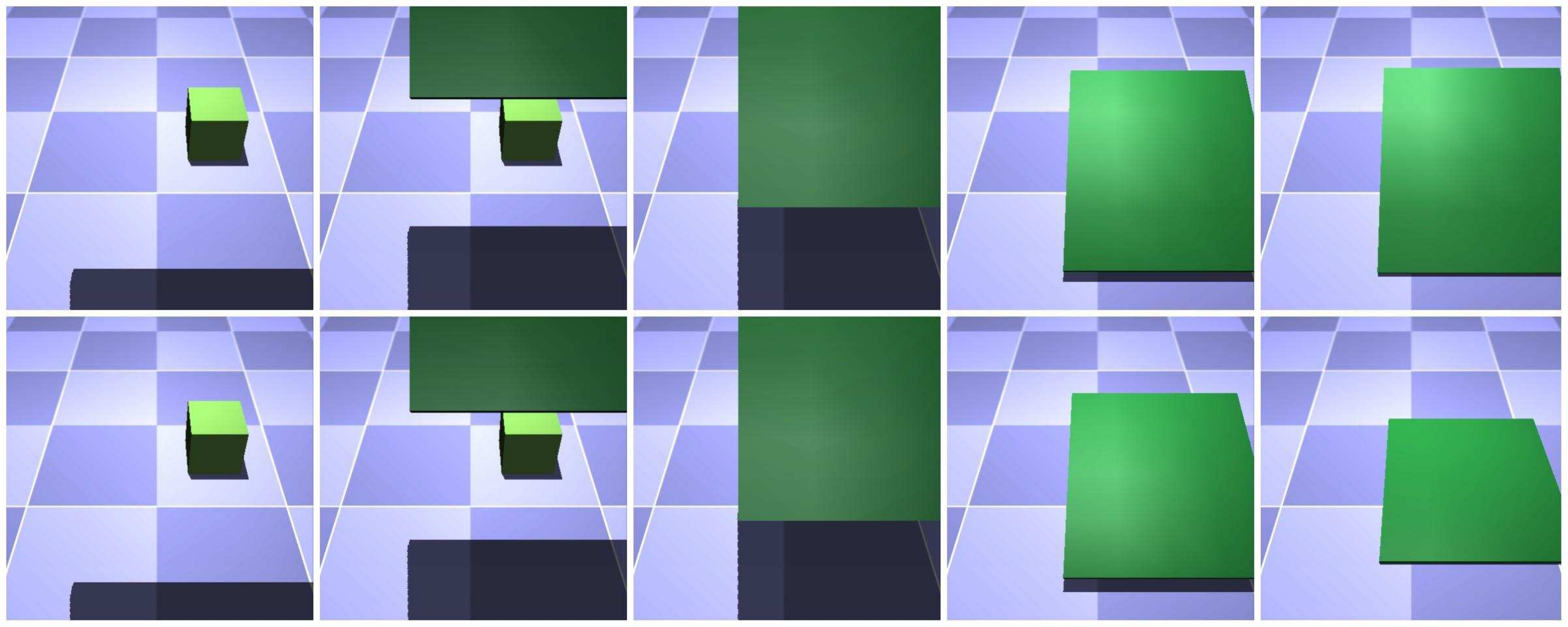}
            \put (2,32) {\huge$\mathbf{a}$}
        \end{overpic}
    \end{subfigure}%
    \vspace{0.15cm}
    
    \begin{subfigure}[]{\linewidth}
        \centering
        \begin{overpic}[width=\textwidth]{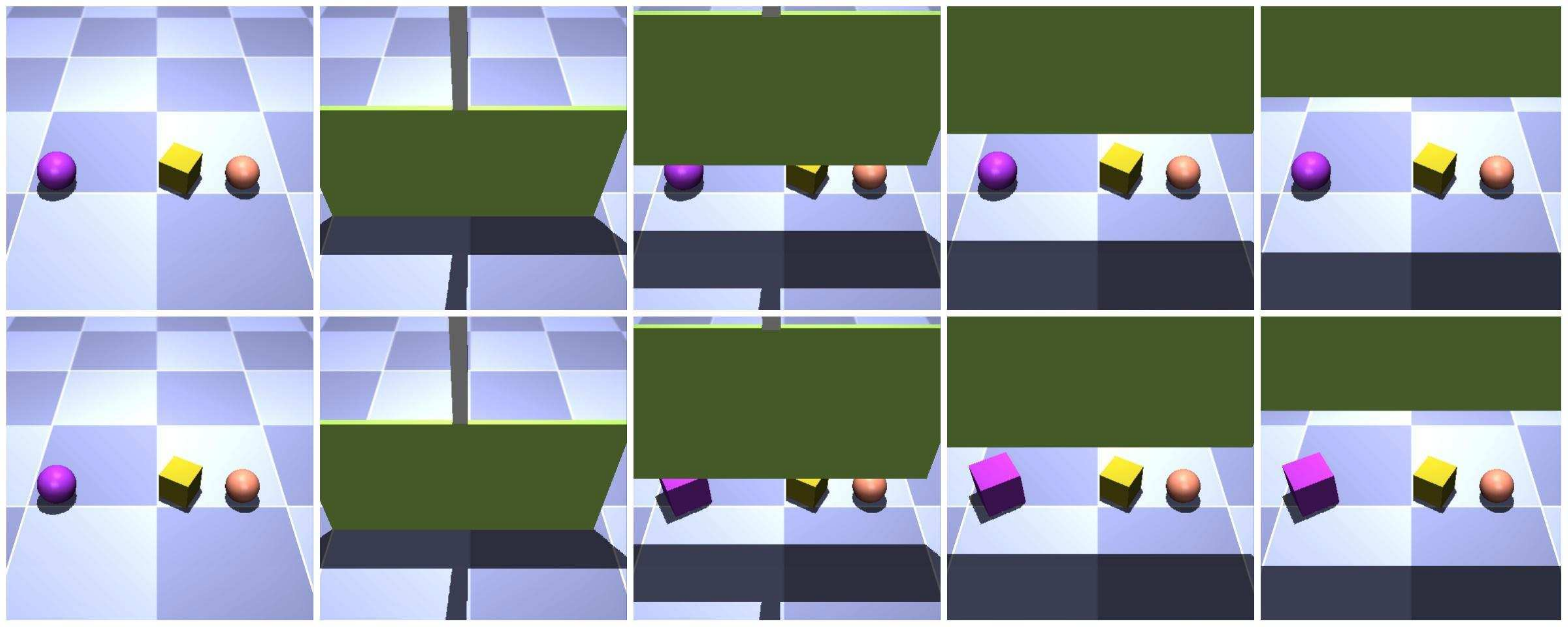}
            \put (2,32) {\huge$\mathbf{b}$}
        \end{overpic}        
    \end{subfigure}
    \vspace{0.15cm}

    \begin{subfigure}[]{\linewidth}
        \centering
        \begin{overpic}[width=\textwidth]{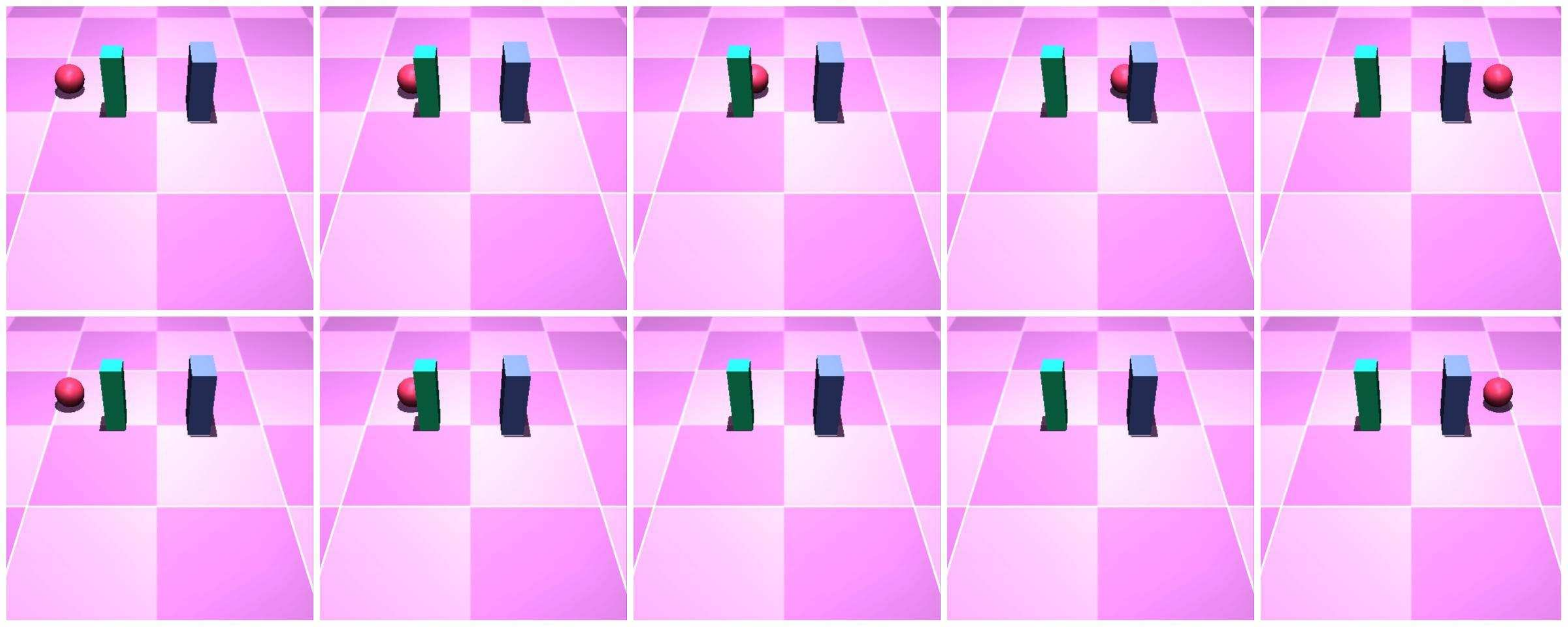}
            \put (2,32) {\huge$\mathbf{c}$}
        \end{overpic}        
    \end{subfigure}
    \vspace{0.15cm}
    
    \begin{subfigure}[]{\linewidth}
        \centering
        \begin{overpic}[width=\textwidth]{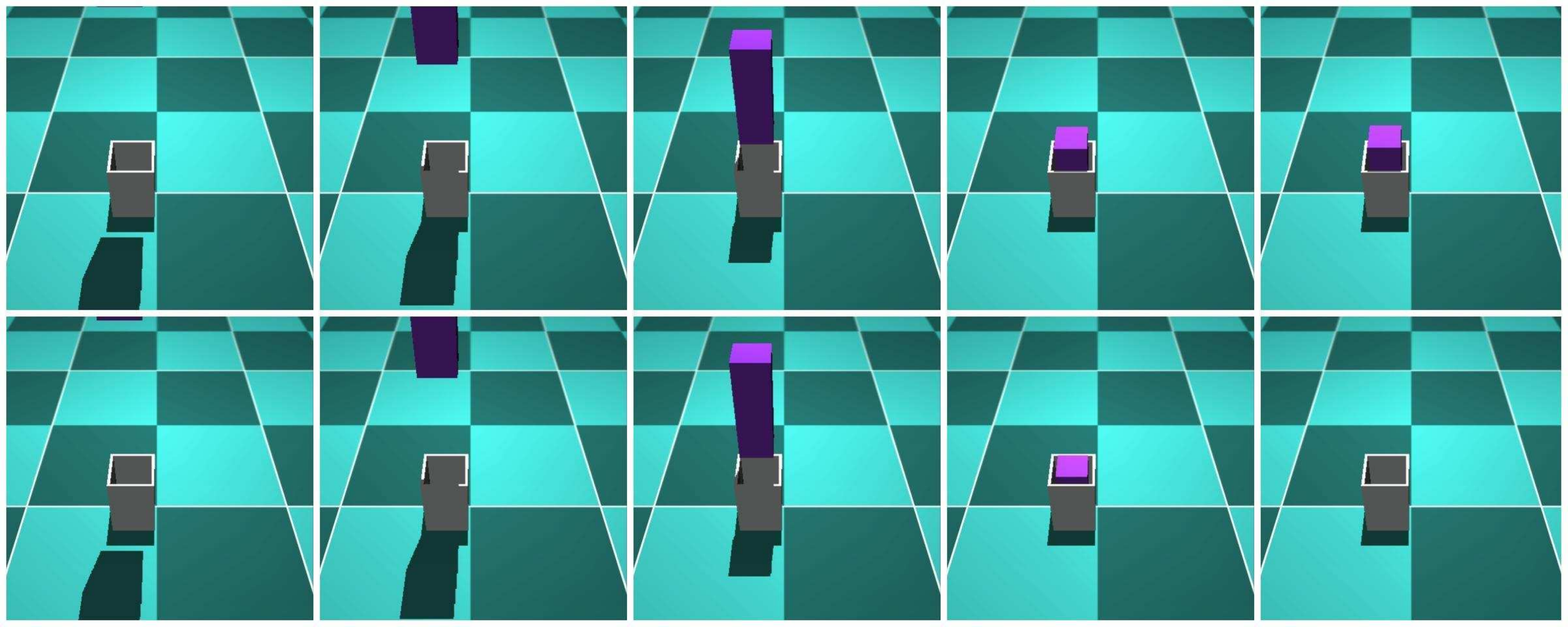}
            \put (2,32) {\huge$\mathbf{d}$}
        \end{overpic}                
    \end{subfigure}
    \vspace{0.15cm}
    
    \begin{subfigure}[]{\linewidth}
        \centering
        \begin{overpic}[width=\textwidth]{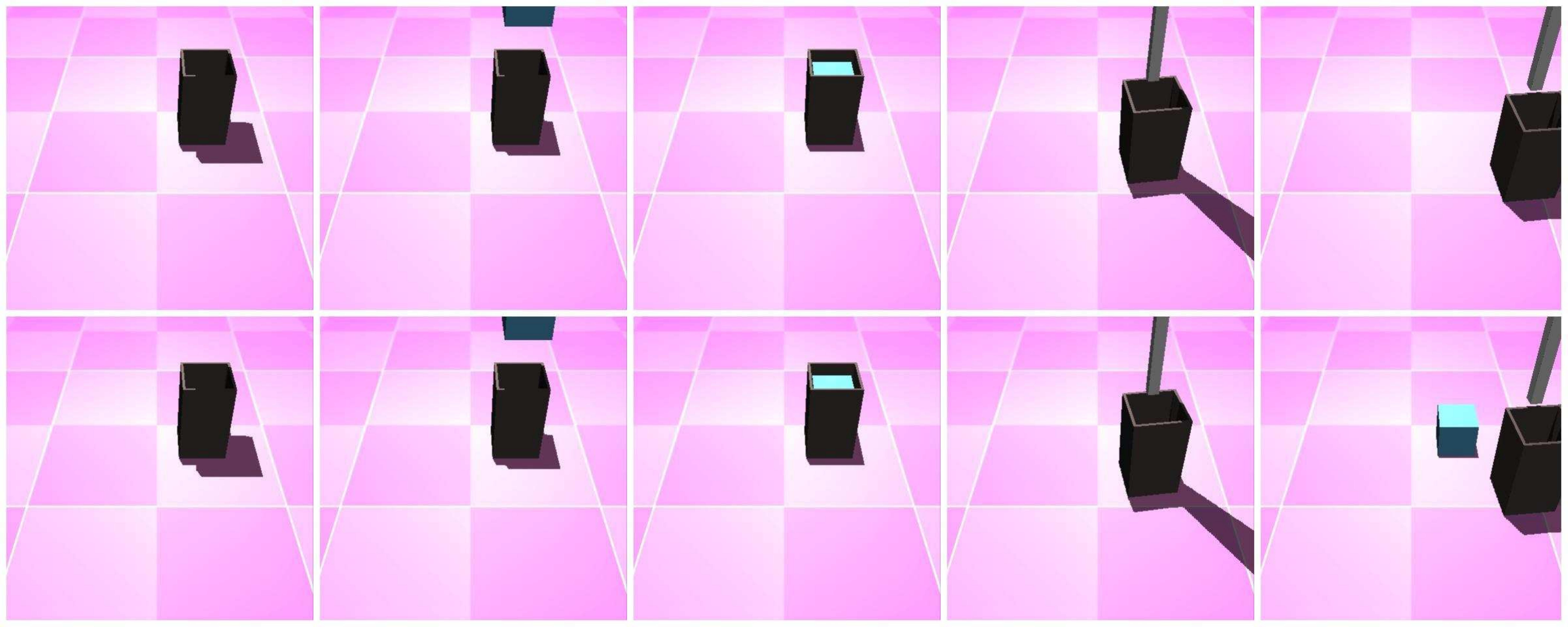}
            \put (2,32) {\huge$\mathbf{e}$}
        \end{overpic}        
    \end{subfigure}
    \caption{Example (subsampled) test probes for each dataset. The first and second rows show consistent and inconsistent probes, respectively. Categories: $a$, Object Persistence; $b$, Unchangeableness; $c$, Continuity; $d$, Solidity; $e$, Containment}
    \label{fig:test_data}
\end{figure}

\begin{figure}[t!]
    \centering
    \begin{subfigure}[]{\linewidth}
        \centering
        \begin{overpic}[width=\textwidth]{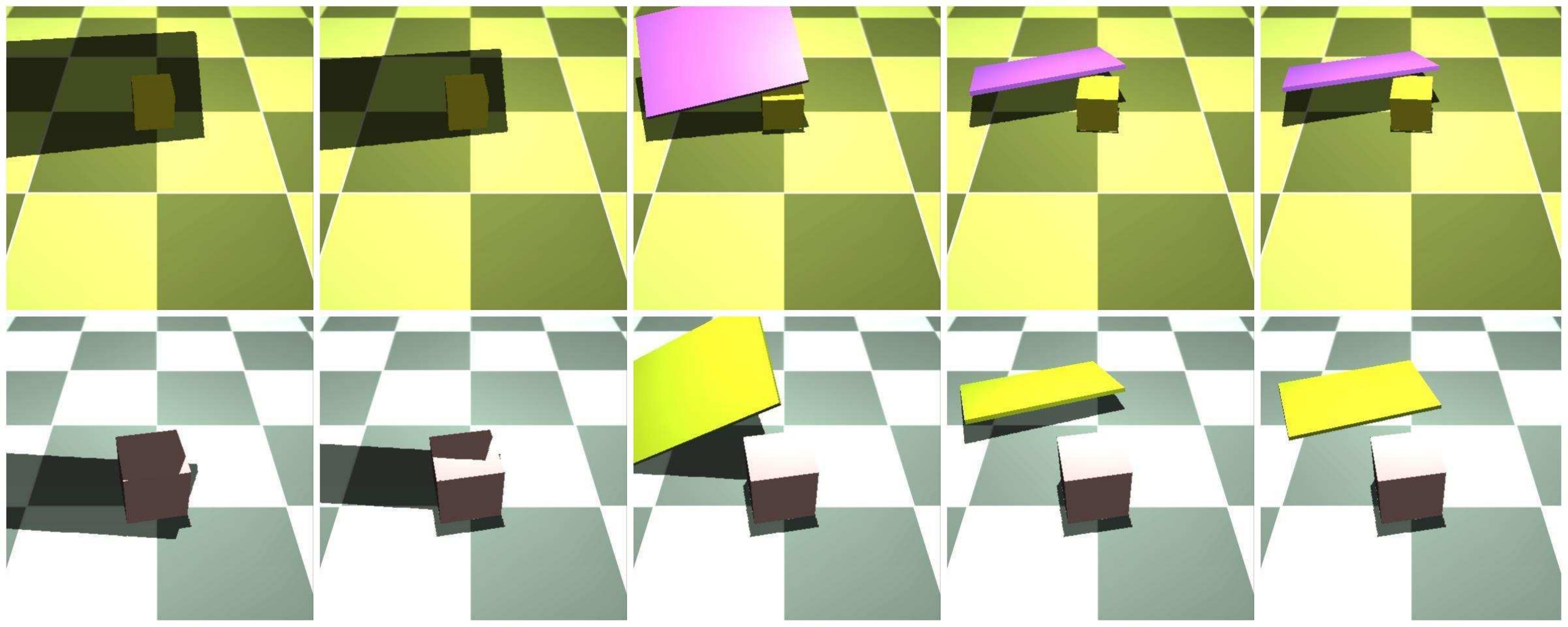}
            \put (2,32) {\color{white}\huge$\mathbf{a}$}
        \end{overpic}
    \end{subfigure}%
    \vspace{0.15cm}
    
    \begin{subfigure}[]{\linewidth}
        \centering
        \begin{overpic}[width=\textwidth]{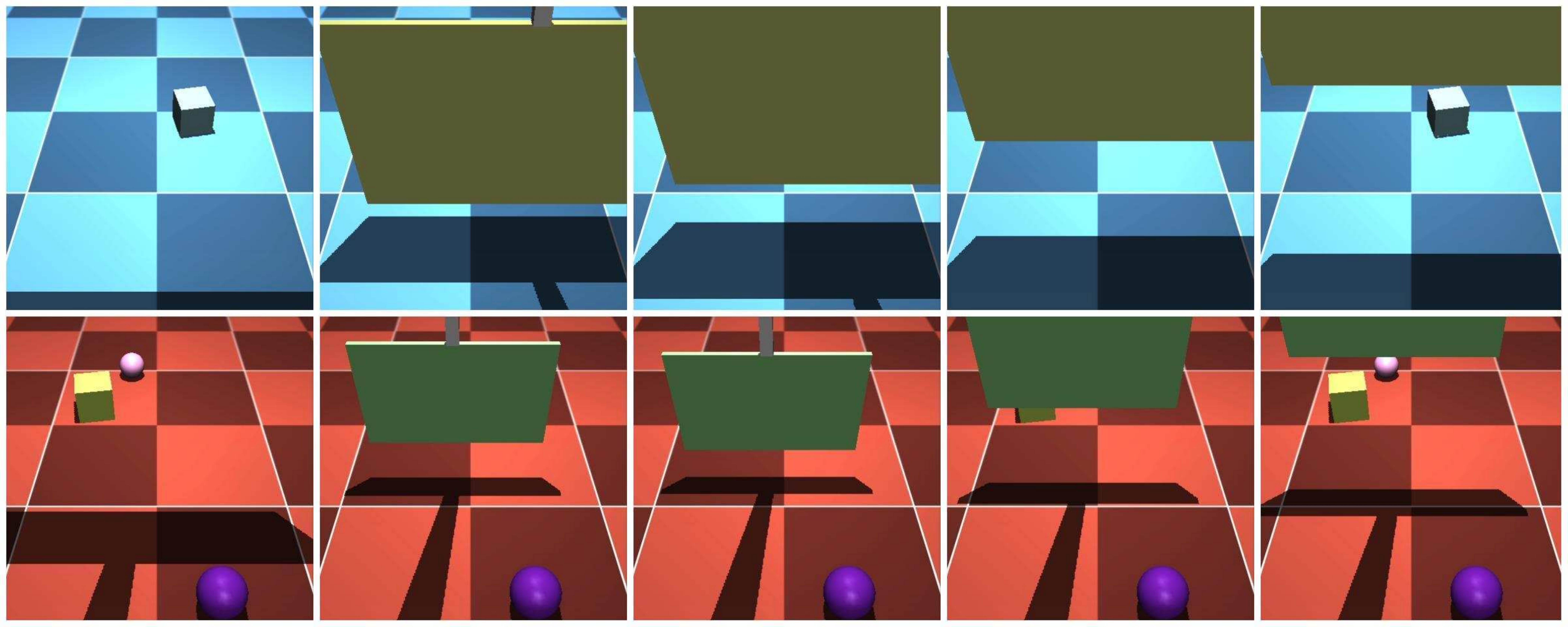}
            \put (2,32) {\huge$\mathbf{b}$}
        \end{overpic}        
    \end{subfigure}
    \vspace{0.15cm}

    \begin{subfigure}[]{\linewidth}
        \centering
        \begin{overpic}[width=\textwidth]{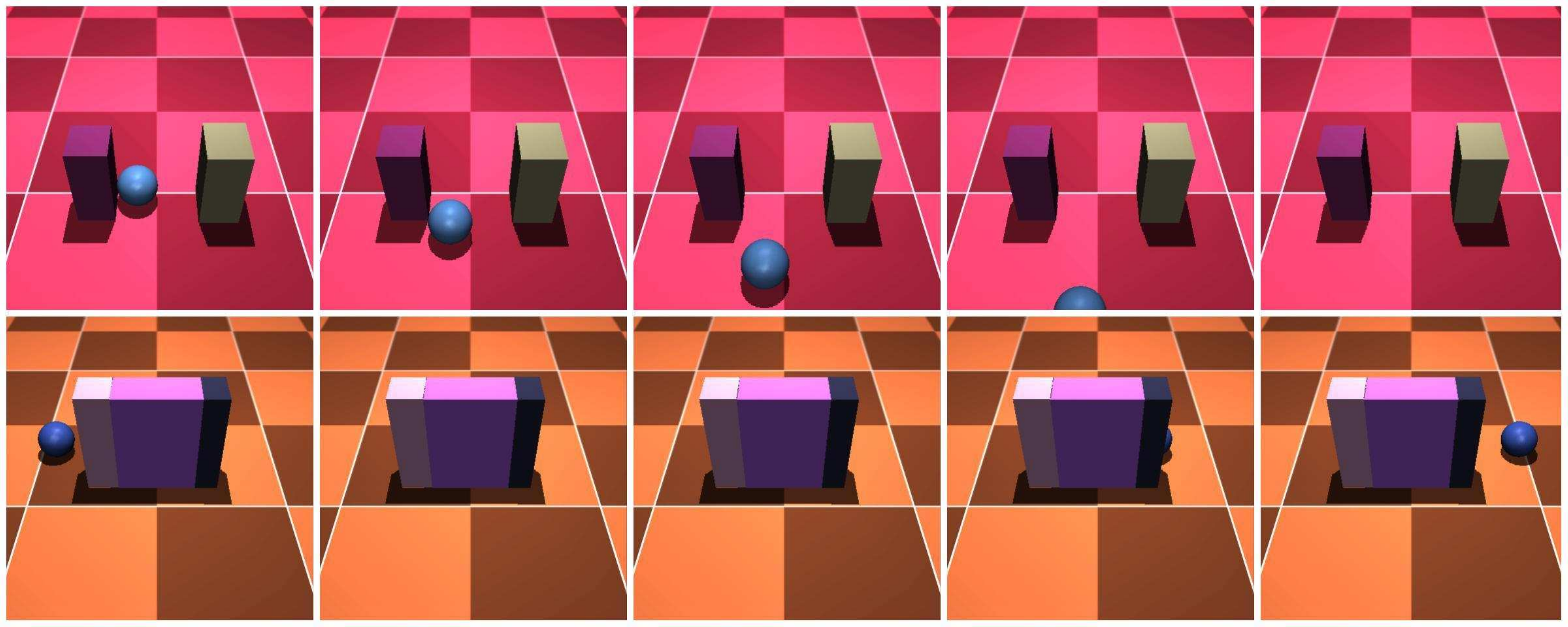}
            \put (2,32) {\huge$\mathbf{c}$}
        \end{overpic}        
    \end{subfigure}
    \vspace{0.15cm}
    
    \begin{subfigure}[]{\linewidth}
        \centering
        \begin{overpic}[width=\textwidth]{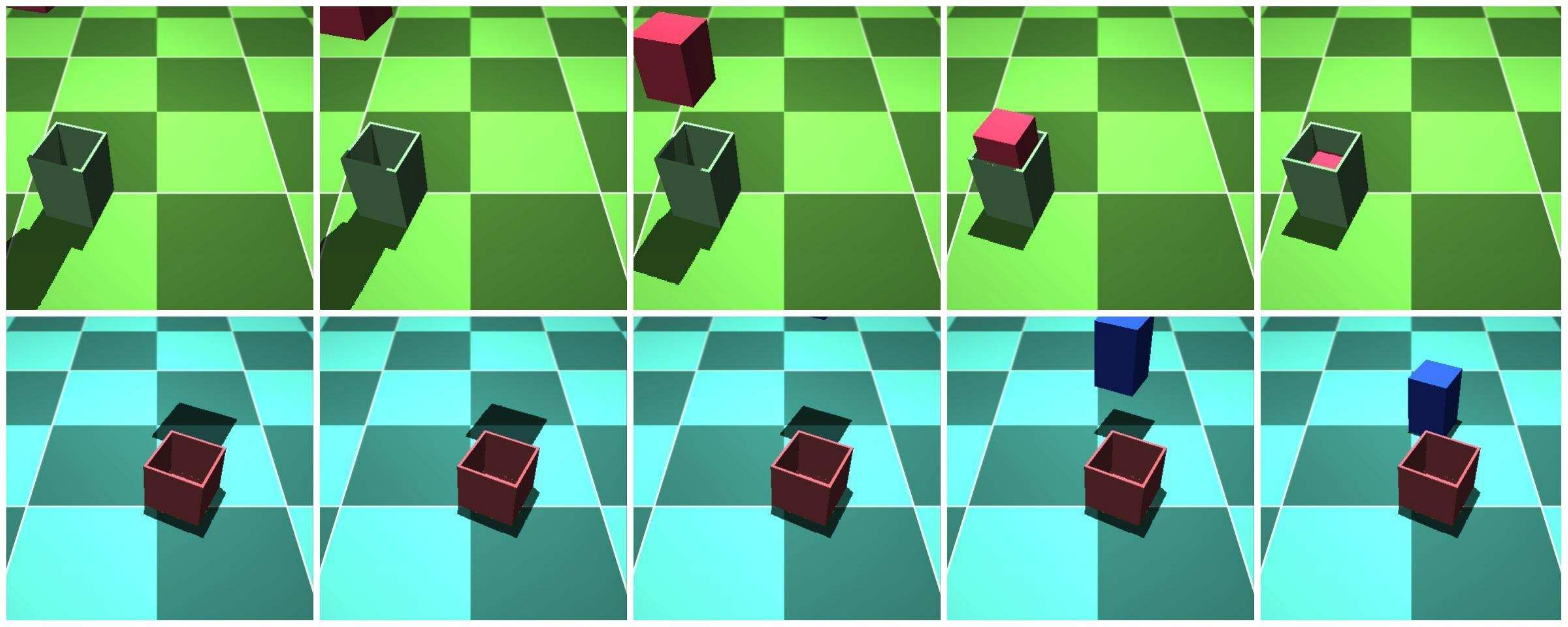}
            \put (2,32) {\huge$\mathbf{d}$}
        \end{overpic}                
    \end{subfigure}
    \vspace{0.15cm}
    
    \begin{subfigure}[]{\linewidth}
        \centering
        \begin{overpic}[width=\textwidth]{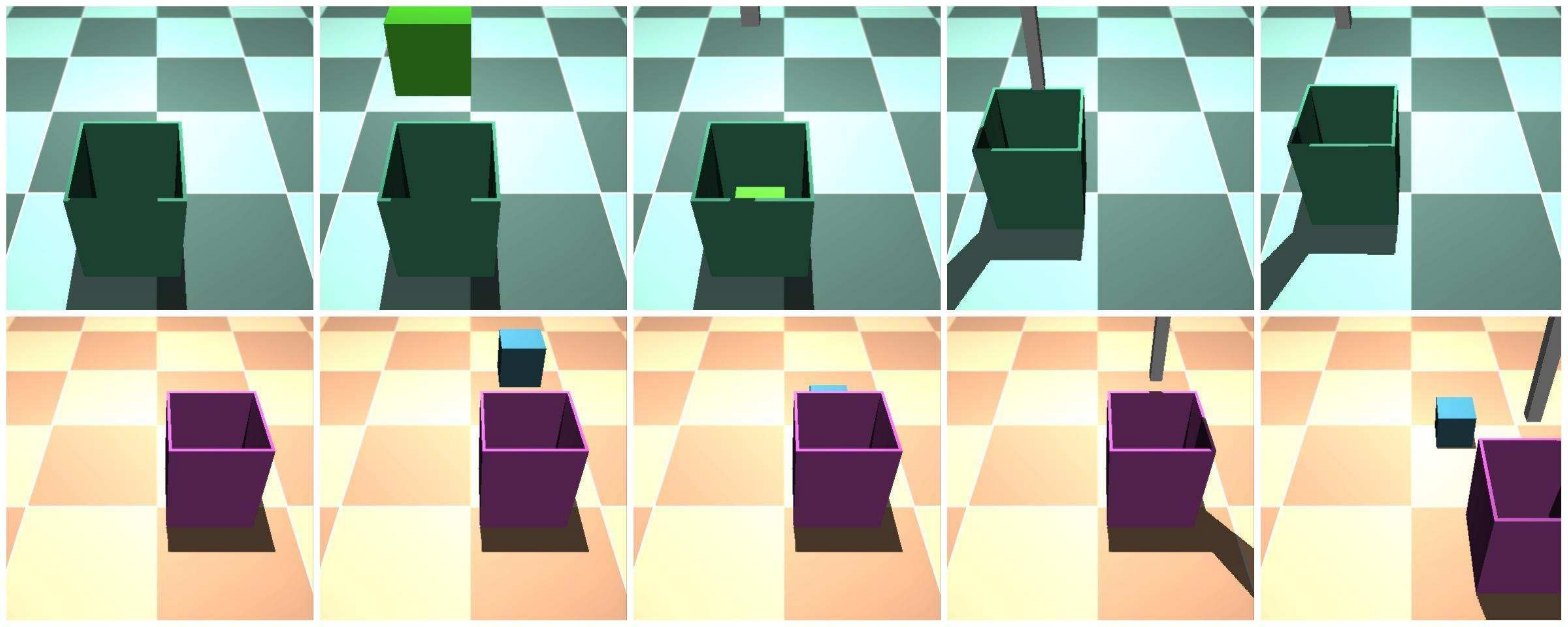}
            \put (2,32) {\huge$\mathbf{e}$}
        \end{overpic}        
    \end{subfigure}
    \caption{Examples of (subsampled) training data (consistent examples and/or controls). Categories: $a$, Object Persistence; $b$, Unchangeableness; $c$, Continuity; $d$, Solidity; $e$, Containment}
    \label{fig:train_data}
\end{figure}

Finally, we look at containment events, which requires understanding that an object inside a container remains inside as the container is moved.  As is the case with earlier studies in developmental psychology~\cite{Hespos2001}, probes of this dataset involve an object falling inside a container.  Once the object (a cube) falls to the bottom of the container it is occluded by the container's walls, and the container is then moved by a rod that descends from the top of the scene (Figure~\ref{fig:test_data}$e$, top row).  In the inconsistent probes (Figure~\ref{fig:test_data}$e$, bottom row) the movement of the container reveals the cube to have remained in the same position on the floor (where it and the container previously were), and consistent probes have the cube moving inside the container as expected.  Consistent examples (Figure~\ref{fig:train_data}$e$, top row) are like consistent probes, but container position and size are not constrained to occlude the cube.  In controls (Figure~\ref{fig:train_data}$e$, bottom row) the cube may fall behind the container (therefore also becoming occluded), and when the container is moved the cube remains where it was (similar to inconsistent probes).  The cube may also fall at a completely randomized position before the container is moved.

\section{Model}
\label{model}

In order to make an initial test of the VOE method, we require a model that has some form of memory, and readily allows for computation of the KL-divergence.  Alhtough we could have selected from a number of models that meet this criteria, we chose a previously proposed generative temporal model with external memory~\cite{Gemici2017}, as it showed good initial performance.  This takes the form of a variational recurrent neural network
(VRNN) with the Least-Recently Used (LRU) memory mechanism for 
memory.  This is most easily understood as a recurrent neural network where each
time step or ``core'' is a variational autoencoder (VAE) with the hidden
state determined from the external memory.   Unless where stated otherwise, 
implementation details are exactly as in the LRU model of \citet{Gemici2017}.  One important
deviation from standard variational autoencoders is that in the case of VRNNs,
the prior distribution is generated at each time step as a function of the
previous observations in the sequence.


\tikzstyle{block} = [rectangle, draw, text centered]
\tikzstyle{line} = [draw, -latex']
\tikzstyle{cloud} = [draw ]

\tikzstyle{stochastic} = [draw, ellipse ]

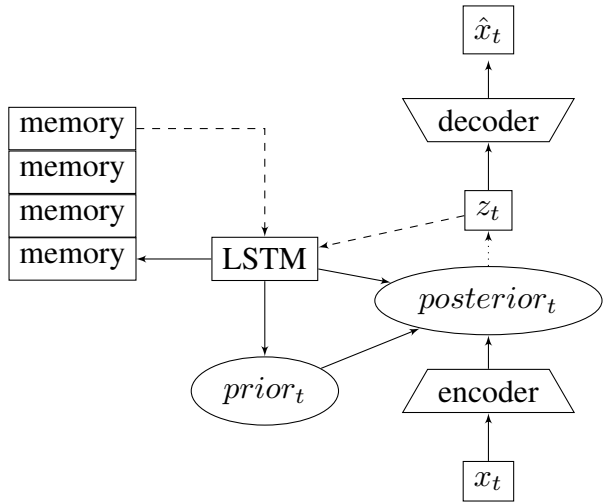
\begin{figure}[ht]
\centering
\begin{tikzpicture}[auto,scale=1.2, every node/.style={scale=1.2}]
	\node [block, outer sep=2pt] (recon) { $\hat{x}_t$ };
	\node [trapezium, trapezium angle=-60, draw, below of=recon] (decoder) { decoder };
	\node [rectangle, draw, below of=decoder] (z) {$ z_t $};
	\node [stochastic, below of=z] (posterior) { $ posterior_t$ };
	\node [trapezium, trapezium angle=60, draw, below of=posterior] (encoder) { encoder };
	\node [block, below of=encoder] (input) { $x_t$ };
	\node [stochastic, left = of encoder] (prior) { $ prior_t$ };
	\node [rectangle, draw, above = of prior] (lstm) { LSTM };
	\node [rectangle, draw, outer sep=0pt, left = of lstm] (memory_base) { memory };
	\node [rectangle, draw,anchor=south] at (memory_base.north) (memory_prev1) { memory };
	\node [rectangle, draw,anchor=south] at (memory_prev1.north) (memory_prev2) { memory };
	\node [rectangle, draw,anchor=south] at (memory_prev2.north) (memory_top) { memory };

    \path [line] (input) -- (encoder);
    \path [line] (prior) -- (posterior);
    \path [line] (encoder) -- (posterior);
    \path [line, dotted] (posterior) -- (z);
    \path [line] (z) -- (decoder);
    \path [line] (decoder) -- (recon);

    \path [line, dashed] (z) -- (lstm);
     \path [line] (lstm) -- (memory_base); 
    \path [line] (lstm) -- (prior); 
    \path [line] (lstm) -- (posterior); 
     \path [line, dashed] (memory_top) -| (lstm);
\end{tikzpicture}
\caption{Model architecture for a single time step.  Dotted lines depict sampling from a distribution and dashed lines denote a one time-step delay.  Thus, the sampled $z_t$ and the memory state from the \emph{previous} time step are used to calculate the prior and posterior for the current time step. }
\label{fig:model}
\end{figure}

The model operates on a sequence of
observations, $x_{\leq T}={\{x_1,x_2,\cdots, x_T\}}$ and approximately infers a
set of latent variables $z_{\leq T}={\{z_1,z_2,\cdots, z_T\}}$.  In order to specify our VRNN, we must specify four maps as defined in~\cite{Gemici2017}.  The \emph{posterior map} defines the approximate posterior distribution over 256 Gaussian latents.  This is calculated by adding to the prior the output of a multi-layered perceptron (MLP) with 2 layers of 512 units using the $tanh$ activation function clipped between $[-10, 10]$.  The MLP takes as input an image passed through a convolutional neural network (described in the Appendix), the memory output determined at the previous timestep, and another copy of the prior for the current timestep. The \emph{prior map} specifies the prior distribution over latents using the same architecture as the posterior map (with its own set of weights, of course).  It depends only on the \emph{history} of latent variables and observations via the external memory's output.  This memory output is the same output fed to the posterior.  The \emph{observation map} specifies the parameters of the likelihood function via a deconvolutional neural network (see Section A in Appendix) as a function of the sample from the posterior distribution.
The \emph{transition map} specifies how the hidden state, $h_t$, and memory, $M_t$, are updated at each time step.  For this we use the LRU memory mechanism with a long short-term memory (LSTM) ~\cite{Hochreiter1997} controller using 128 units to control reads.  Output from the memory is read as the weighted sum over 3 memory slots.  The LSTM takes as input the previous hidden state and the sampled latents inferred for the previous observation.


The external memory has a total of 15 memory slots and 100 units per
slot.  We use a Bernoulli distribution for the likelihood of our observations by scaling pixels to be in the range $[0,1]$.  We also include an additional
term in the objective function to regularize the prior distribution (and by proxy the posterior distribution), towards a unit Gaussian.  This is scaled by $1e^{-4}$ in the objective function.


\subsection{Training and Evaluation}

For each concept-category dataset, we trained on 100,000 sets of consistent examples and controls each. Testing was conducted over 2,000 pairs of probes for a  hyperparameter search, and another 8,000 pairs of probes were used for validation after the hyperparameter sweep was completed.  Each video was restricted to 15 frames at 64x64 resolution in RGB channels.  Training was done under two regimes.  
In the first case, we trained and tested separate models each on a single dataset.
In the second case, we trained on all the datasets together.  In all cases, we used the Adam~\cite{Kingma2015} optimization algorithm with a batch size of 10, trained for 2 million steps, and a learning rate of $5e^{-6}$.  Although the loss function converged to a stable value well before this point, we noticed continual improvement in reconstructions as well as reductions in relative surprise with continued training in some categories (see Figure~\ref{fig:kl_effect_over_training}).

As the probes were evaluated by the model, the KL-divergence was recorded over all frames for both the consistent and inconsistent probes.  Statistical testing for the KL-divergence between the consistent and inconsistent probes was done with a one-tailed, paired t-test with a significance level of $0.05$.

\begin{figure*}[t!]
  \centering
  \includegraphics[width=\textwidth]{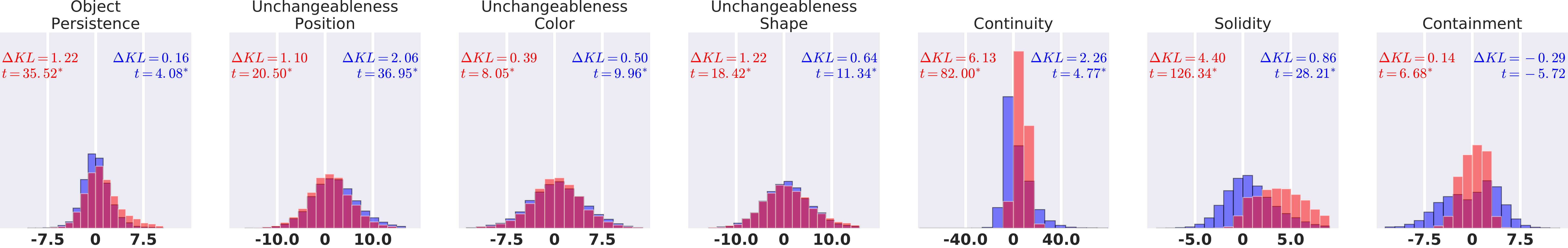}
  \caption{Difference in KL divergence (x-axis) over final evaluation set of 8,000 probe pairs. Histograms show the distribution over the evaluation set with outliers \textgreater 3 standard deviations removed.  Text shows the result of significance tests with an asterisk denoting significance. Red corresponds to models trained separately, blue to model trained concurrently.}\label{fig:combo_histogram_stats_table}
\end{figure*}

\section{Results}

We first evaluated our baseline model trained on each dataset separately.  For all categories we found the surprise for inconsistent probes was higher than consistent probes on average (see Figure \ref{fig:combo_histogram_stats_table}, red text for exact numbers).  This indicates the network does have an ability to recognize violations of these concepts, although the trial-by-trial variability in the histograms in Figure  \ref{fig:combo_histogram_stats_table} indicate there is room for model improvement.

\begin{figure*}[ht]
  \centering
  \includegraphics[width=\textwidth]{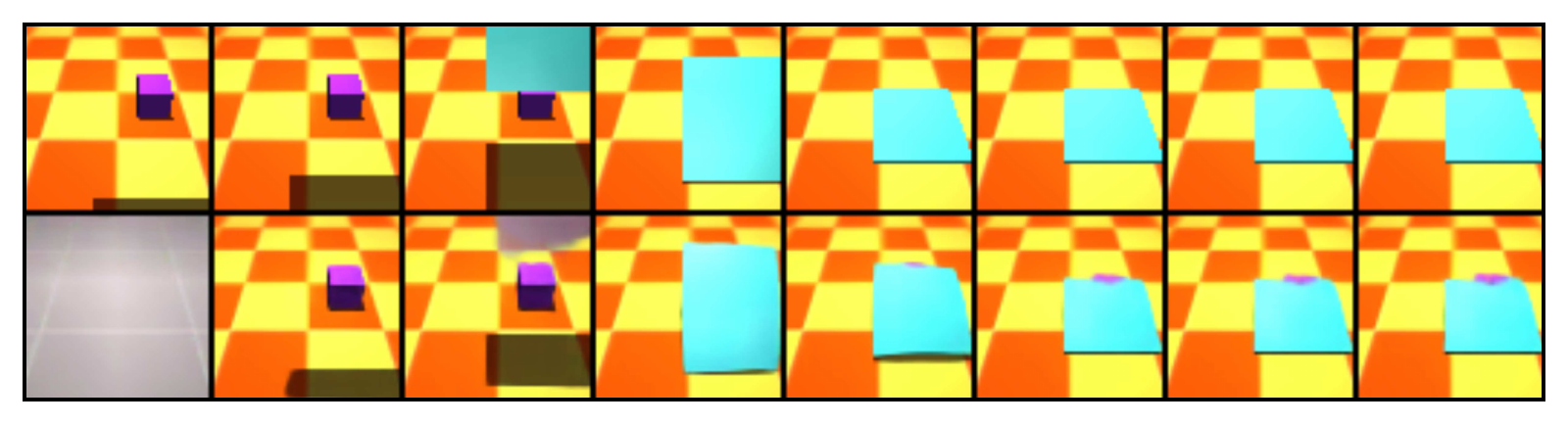}
  \caption{Top row: Subsampled frames from an inconsistent probe for the Object Persistence dataset.  Here physics is violated when the sheet magically collapses the magenta cube.  Bottom row: Network's predictions for the above frame based only on previous frames. Note how it rectifies the input with the correct physics displaying a scene where the sheet rests on top of the magenta cube.}\label{fig:baill_prediction}
\end{figure*}

Although our focus in the present work is on the VOE method, it is worth noting convergent evidence that the network was ``understanding'' scene information in a meaningful way. Visualizations confirm the network attempts to reconcile its knowledge of physics with the violations that occur during inconsistent probes. Figure \ref{fig:baill_prediction} presents an Object Persistence inconsistent probe (top row) along with renderings of the prior per frame (bottom row).  In this video, a magenta cube disappears while occluded by a falling cyan plank.  Here the prior, conditioned on the preceding frames, predicts the sheet should rest on the magenta cube, instead of having the cube disappear.  This is apparent in the portion of the magenta cube protruding beyond the top of the plank in the last four frames. 

While we did not explore generalization beyond the training distribution systematically, initial experimentation indicated that the network was capable of some degree of extrapolation.
We tested the ability of our model to do so by 
introducing a change to the Continuity dataset by introducing probes containing an additional pillar (that never was present during training) between the two pillars (that were present during training).  Crucially, the physical concept at stake remains the same: continuity of a moving object.  We found that the model trained on the Continuity dataset generalized to these new probes (inconsistent probes were significantly more surprising: ($M=3.47;\ t(7999)=5.11,\ p < .05$)), although this generalization did not hold for the jointly trained model ($M=0.17;\ t(7999)=1.58,\ p = .06$).  Additionally, for the model trained only on Unchangeableness, we tested generalization to a maximum of two more objects than present during training.  We found significantly more surprise for inconsistent probes  when we modified position ($M=2.48;\ t(7999)=136.25,\ p < .05$), color ($M=0.98;\ t(7999)=70.44,\ p < .05$), and shape ($M=1.94;\ t(7999)=98.60,\ p < .05$).

\begin{figure}[ht]
  \centering
  \includegraphics[width=\linewidth]{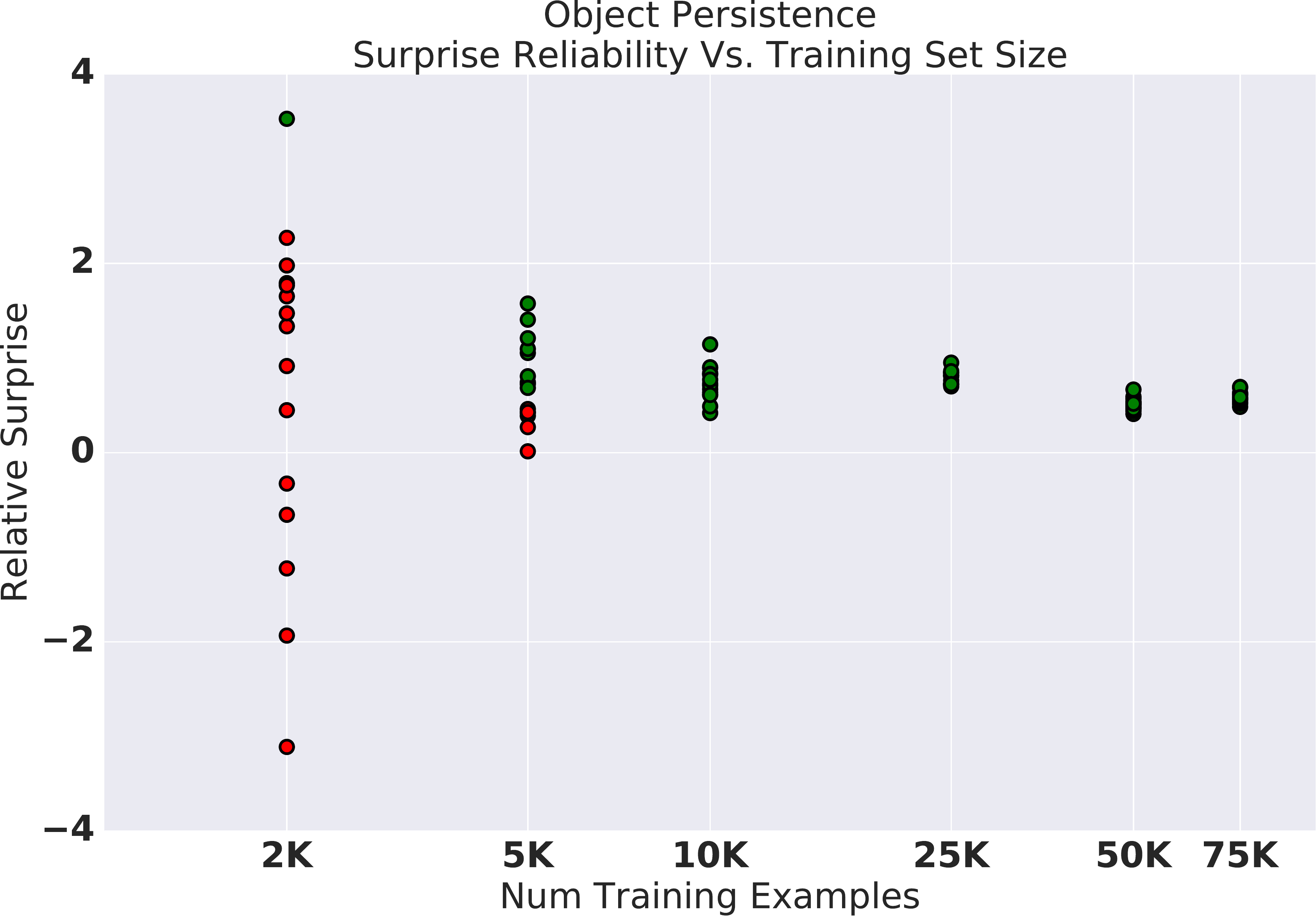}
  \caption{Reliability of Relative Surprise (y-axis) for models trained with different Training Set Sizes (x-axis).  Each point represents a different evaluation set of 2,000 probe pairs with green points showing a statistically significant effect.  As the training set size increases, the variance of the effect size decreases and reliability of statistical significance increases. }\label{fig:effect_size_vs_corpus_size}
\end{figure}

Our initial choice to employ a large training set was aimed at assuring sufficient diversity in the training data. However, we also examined the sensitivity of VOE effects to sample size. Analyses using the Object Persistence dataset revealed significant VOE effects with as few as ten thousand training examples.  As illustrated in Figure \ref{fig:effect_size_vs_corpus_size}, the effect of increasing training-set size was primarily to reduce the variance of VOE effects across independently sampled test sets.

\begin{figure}[ht]
  \centering
  \includegraphics[width=\linewidth]{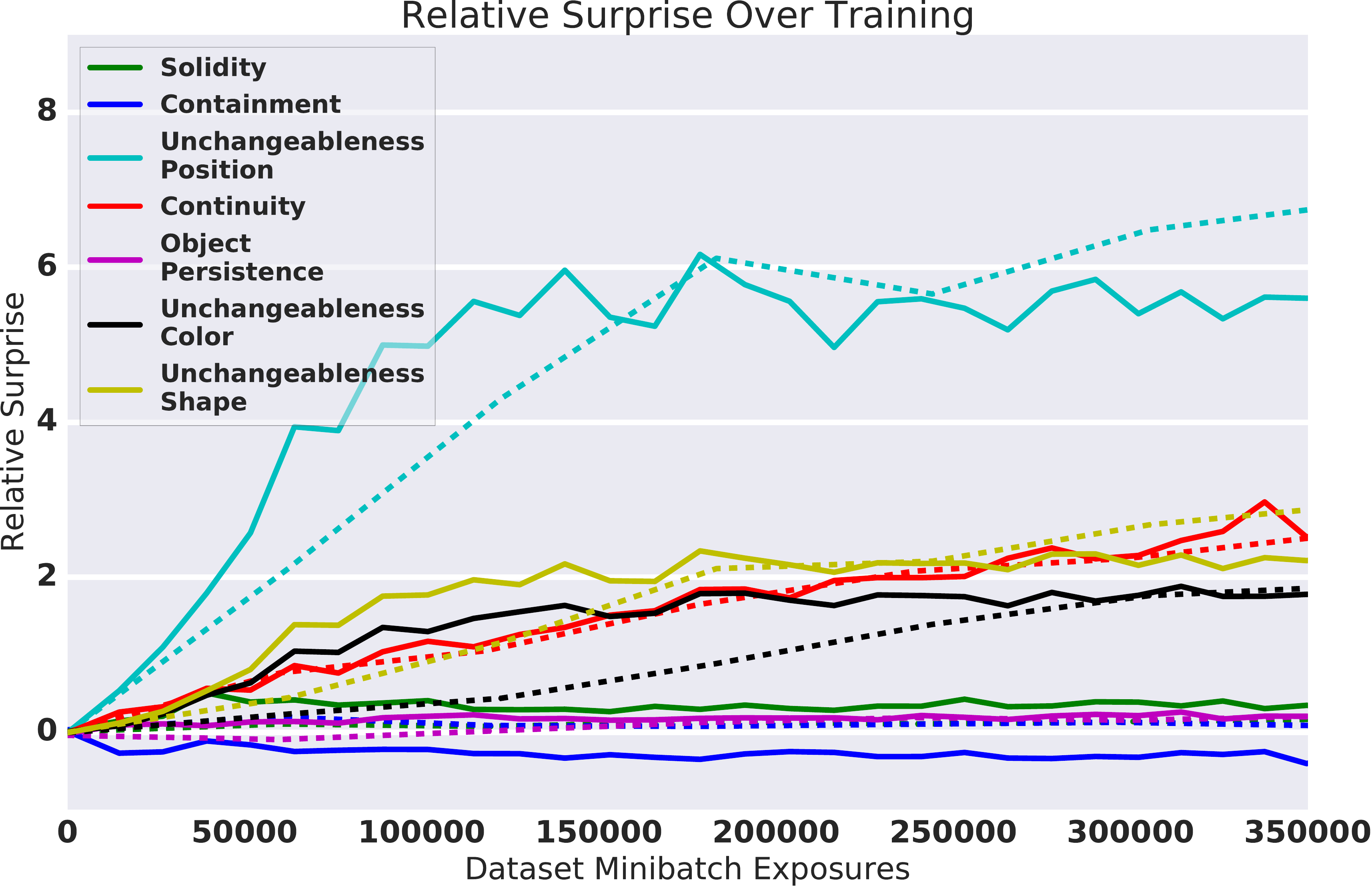}
  \caption{Relative Surprise Over Training for Models Trained on Datasets Individually (dashed lines) or Simultaneously (solid lines)}\label{fig:kl_effect_over_training}
\end{figure}

Next, we examined the results of training the model on all five datasets concurrently, interleaving examples and applying the same evaluation procedure as before. This revealed a statistically significant VOE effect for all categories except Containment (see Figure \ref{fig:combo_histogram_stats_table}). 
By matching the number of training examples per category, we observed that VOE effects emerged more rapidly for most categories during joint training than separate training  (Figure \ref{fig:kl_effect_over_training}). Putting all these results together, it appears that  concurrent training may give rise to both transfer and interference.  

Concurrent with the present work, Riochet and colleagues produced an interestingly related dataset~\cite{Riochet2018}. While not framed specifically in terms of VOE, their ``IntPhys'' dataset is also concerned with the ability of systems to understand intuitive physics and is amenable to our VOE technique.  We found that the same model used in our other experiments also showed a VOE effect when trained and tested on their visually quite different ``dev'' dataset ($M=0.95;\ t(209)=2.14,\ p < .05$). 

\section{Discussion}

Here we summarize our contributions, related work in the literature, and promising future directions.

\subsection{Present Work}
 

The initial experiments presented provide a clear demonstration of the potential utility of the VOE method for artifical intelligence, as pioneered in developmental psychology.  As we have argued, the appeal of the paradigm lies not only in its focus on surprise, but also in the use of carefully controlled probe stimuli, assessing mastery of specific physical principles.  The work presented only scratches the surface of the developmental psychology literature. The latter points to a wide range of phenomena, and provides a rich source of probe designs, which could be leveraged in further work.

Our results are an important step toward developing integrated models for learning intuitive physics, tackling the problem by investigating how far can standard deep learning methods take us in capturing expectations about intuitive physics.
In our experiments, we employed a relatively generic model (our only requirements were the ability to compute the KL-divergence and some form of memory) as a baseline for testing. This memory-augmented VRNN displayed a surprising capacity to assimilate basic physical concepts. Whereas it has recently been asserted that ``Genuine intuitive physics...is unlikely to emerge from gradient-based learning in a relatively generic neural network''~\cite{Lake2017}, the results we have reported suggest it may be important to give this a try, if only to determine where the approach truly breaks down. The present experiments suggest that a fairly generic neural network architecture can deal with significant train-test separation, as well as considerable visual and dynamic diversity, to extract fundamental physical principles. Some potential ways of testing the limits of such successes are noted below. 

\subsection{Related Work}
\label{related_work}


There is a rapidly growing literature on prediction and learning in physics domains (see~\citet{Byravan2017} for a review). The aspect of this literature most relevant to the present work involves the techniques that have been used to assess physics knowledge in AI systems.  \citet{Battaglia2013} used a ready-made physics engine as a forward model to perform a number of classification and regression tasks related to the dynamics of a tower of falling blocks.  \citet{Lerer2016} also study block towers, and use a bottom-up learning approach to classify stability and predict object masks for falling blocks from pixels, but regresses against ground truth state information.  It is also not a true forward model in that it only makes predictions from the initial state.  The approach taken here generalizes this by looking at full step-by-step prediction. 

Interaction networks~\cite{Battaglia2016} and the neural physics engine~\cite{Chang2017} make architectural modifications designed to promote object- and relation- centric representations in order to predict dynamics. By focusing on the accuracy of step-by-step predictions, they may be complementary to our VOE approach.

Beyond the intuitive physics literature, the work we have presented links with other recent proposals to apply tools from experimental psychology to analyze AI systems \cite{Lake2015, Ritter2017, Leibo2018}.

\subsection{Future Work}
\label{future_work}




This work presents a small-scale initial application of the VOE method, which provides a foundation for a number of next steps.  First, it will be of interest to probe a wider range of concepts, by continuing to draw on the developmental psychology literature related to other concepts such as collisions, number, support, and others~\cite{Baillargeon2012}.  Second, it will be of interest to further separate our train and test sets, ideally moving to training with more free-form, naturalistic data~\cite{Kay2017}, or even video data from infant studies~\cite{Clerkin2017,Smith2017}. As the enterprise expands in this way, we anticipate that more richly structured learning systems may be required, incoprorating such ingredients as attention~\cite{Vaswani2017}, relation networks~\cite{Santoro2017}, attentive comparators~\cite{Shyam2017}, among others.  


We measured surprise as the KL-divergence between the posterior and prior distributions due to its established link to human gaze. However, the possibility for other metrics (e.g. Wasserstein distance) remains.

In order to gain a truly satisfactory knowledge of physics it may also ultimately be necessary to step beyond passive learning to active engagement with the environment, through manipulation and movements that shift the point of view. 

Applications of the VOE method in AI work may also hold the potential to feed insights back to developmental psychology.  Developmental psychologists are still debating various algorithmic-level explanations~\cite{Marr1982} underpinning intuitive physics in infants (see the Physical Reasoning System ~\cite{Baillargeon2012} for an example).   Our framework affords the option to implement and compare these different algorithms and determine which are most successful.

\bibliography{punn.bib} \bibliographystyle{icml2017}


\end{document}


\twocolumn[ \icmltitle{Probing Physics Knowledge Using Tools from Developmental Psychology}]


\appendix
\section {Convolutional Architecture}
For encoding the images, we used 6 ResNet blocks ~\cite{He2016}.  Each block had three 2D convolutions with kernel sizes = 1x1, 3x3, 1x1.  An activation function, rectified linear (ReLU), was only applied after all of the convolutions.  The number of channels in the first two convolutions was 64 while the final convolution had 128 output channels.  For all convolutions, we used a stride of 2x1.  After the series of ResNet blocks, we passed the output through a final MLP with 500 hidden units using a ReLU activation function.

The decoder transposed the encoder architecture to scale from the latent output size to the full observation size.  However, instead of doing a deconvolution to increase the dimensionality, we used a size-preserving convolution followed by an upsampling operation.

\section{Model Representations For Consistent Vs. Inconsistent Probes}

\begin{figure}[H]
      \begin{subfigure}{0.45\linewidth}
        \includegraphics[width=\textwidth]{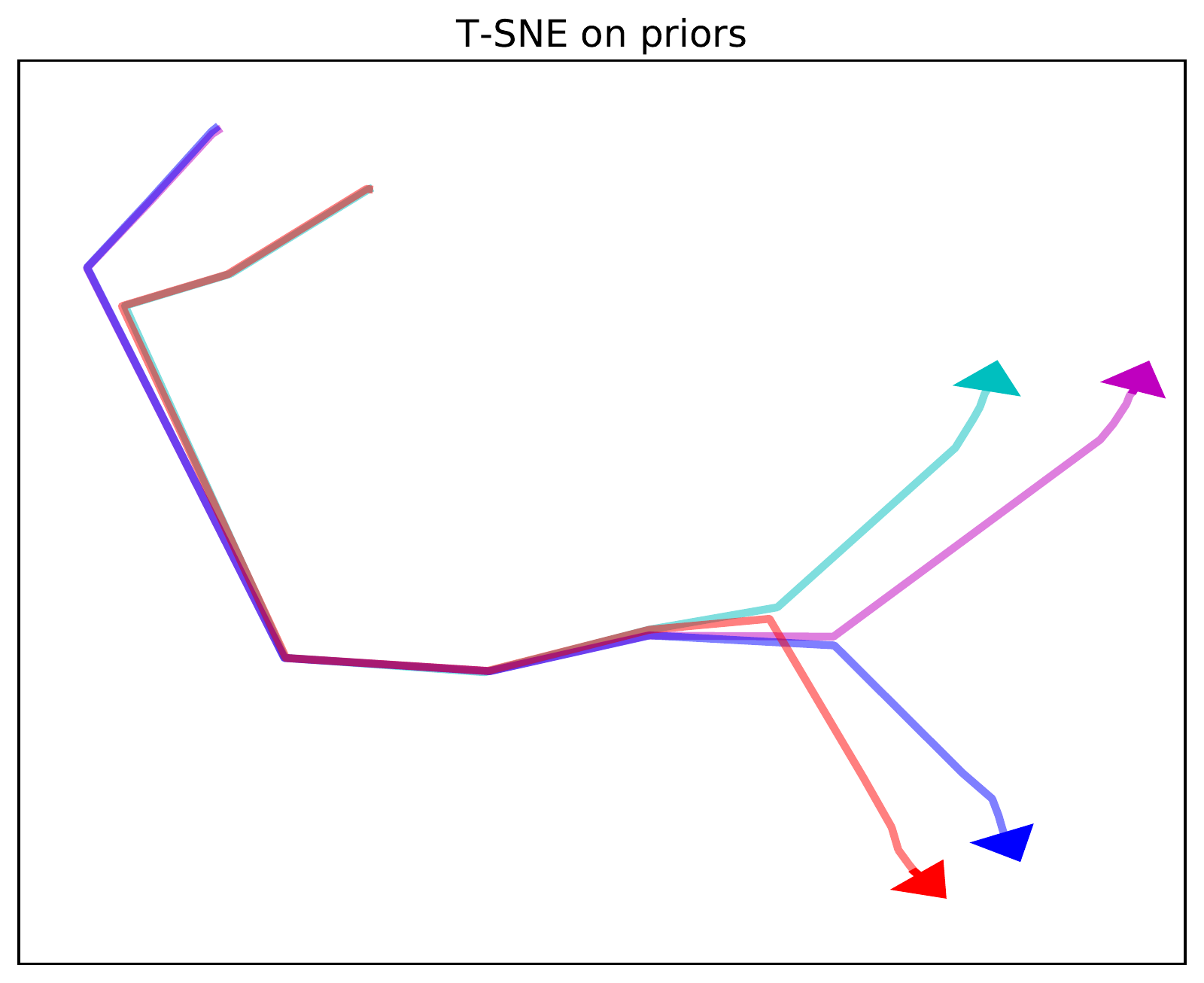}
          \caption{Priors}
          \label{fig:tsne_priors}
      \end{subfigure}
      \hfill
    \begin{subfigure}{0.45\linewidth}
        \includegraphics[width=\textwidth]{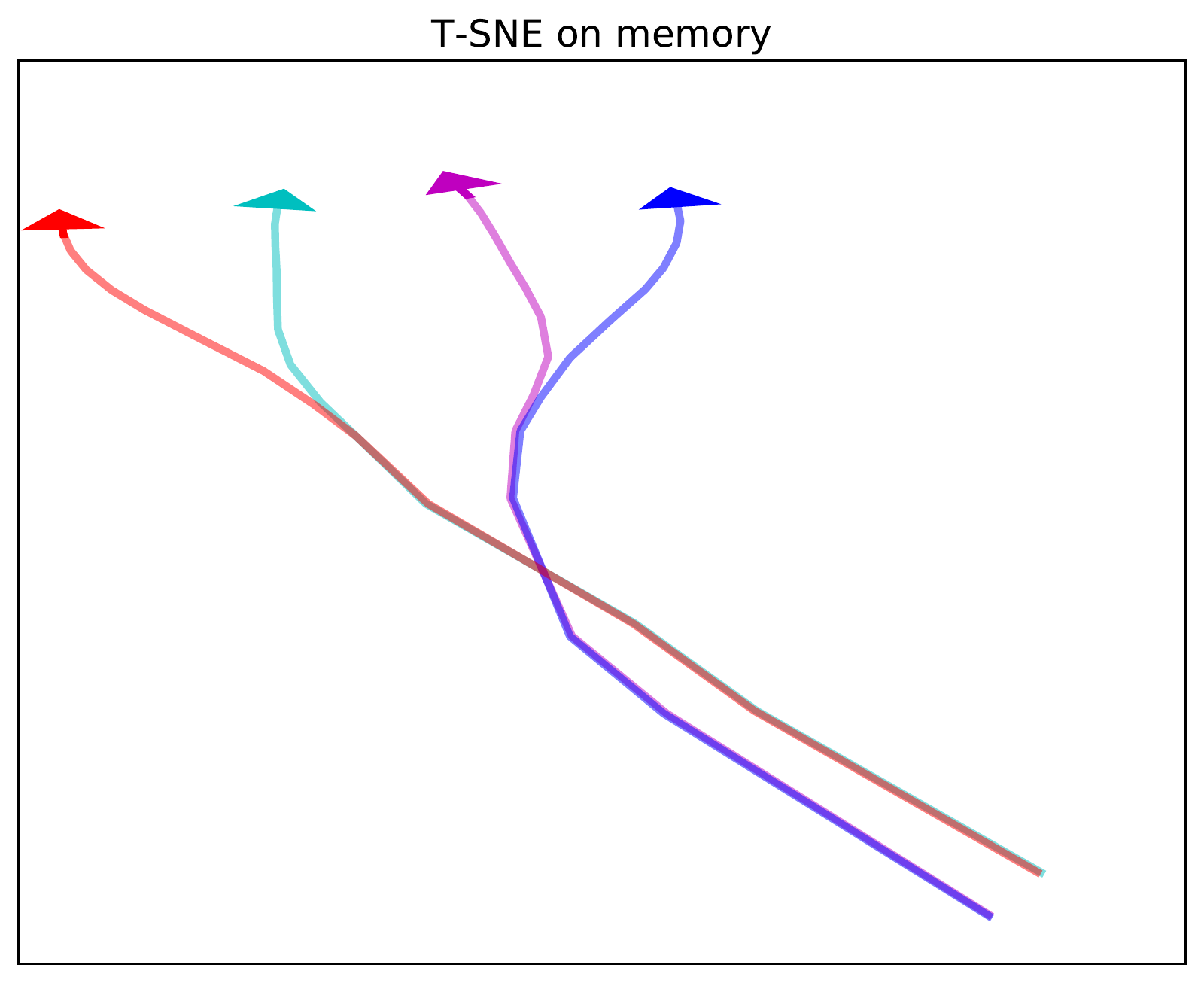}
          \caption{Memory}
          \label{fig:tsne_memory}
      \end{subfigure}
      
  \caption{t-SNE Embeddings}
  \label{fig:tsne}
\end{figure}

In Figure~\ref{fig:tsne}, we show a t-SNE ~\cite{Van2008} representation of the time courses of priors (Figure~\ref{fig:tsne_priors}) and memory (Figure~\ref{fig:tsne_memory}) for counter-balanced probes in the Object Persistence category.
Counter-balanced probes make up 2 pairs of consistent and inconsistent probes, all of which can start and end in only two possible ways.  In these plots, the consistent probes are represented by blue and cyan curves, and the inconsistent probes are represented by red and magenta curves.  In the prior plot, we can see that initially the red and cyan curves are paired correctly (because they begin identically), as are the blue and magenta.  Later in the sequence (moving from the base of the curve in the direction of the arrowhead) this matching is reversed in a manner consistent with the change induced in the inconsistent probes.  Due to the manipulation, the videos represented by red and blue end identically, and likewise for cyan and magenta.  When examining memory, we see the same correct initial pairing of colors as with the priors.  On the other hand, when the manipulation is introduced we do not see a reversal of the pairing (as we do with the priors), but rather a forking effect.  This indicates that the memory maintains information to correctly keep each trajectory distinct based on its entire history, while still able to use the appropriate parts of memory when needed for accurate prior estimation.

\clearpage
\onecolumn

\section{Model Predictions for Inconsistent Probes} \label{section:model_predictions}
Below we show more examples of the model's predictions for inconsistent probes that include impossible manipulations. Frames are subsampled compared to the actual input received by the model.

\begin{figure}[H]
    \centering
        \includegraphics{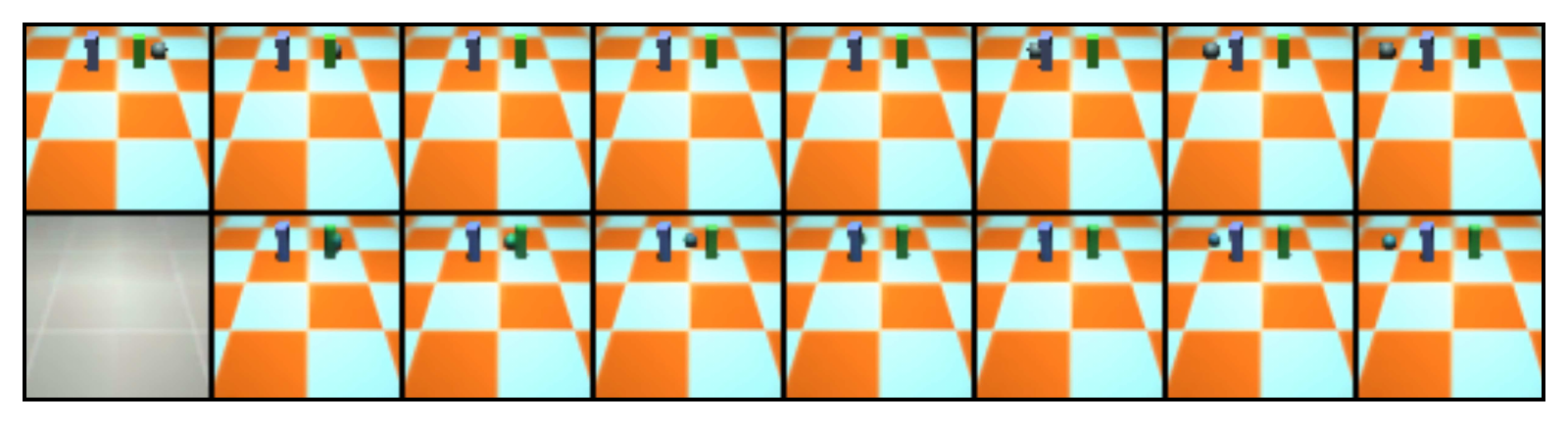}
        \caption{) Network predictions (bottom row) for Continuity inconsistent probe (top row).  Even though the inconsistent probe has the ball disappear between the pillars, the network predicts it should appear.}                  
        \label{fig:continuity_predictions}
\end{figure}
    
\begin{figure}[H]
        \centering
        \includegraphics{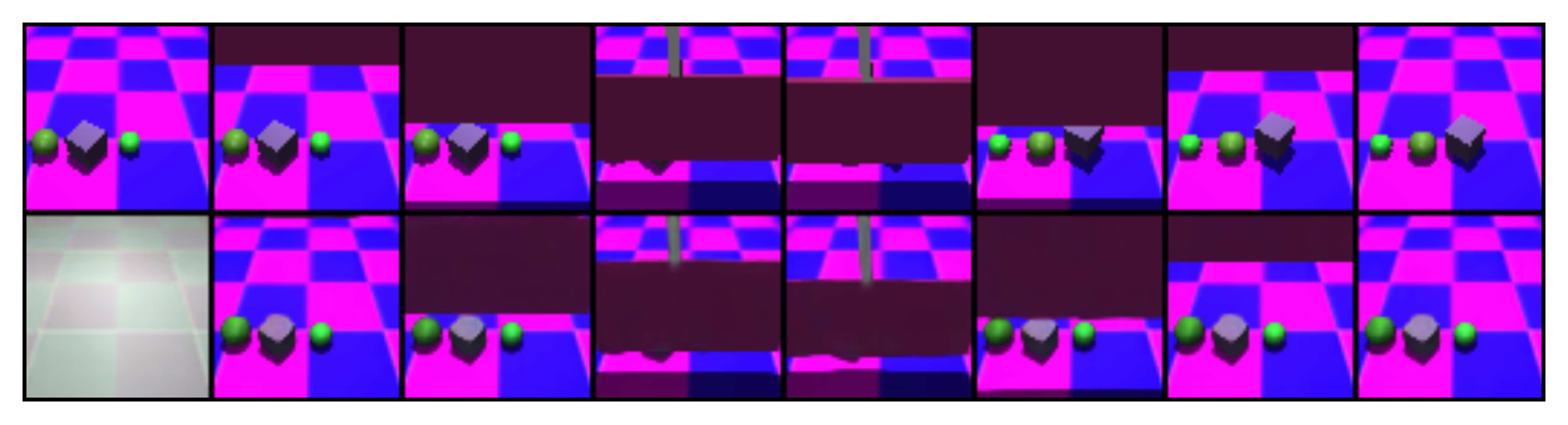}
        \caption{) Network predictions (bottom row) for Unchangeableness Position inconsistent probe (top row).  Even though the object positions have been swapped after the occluder is lifted, the network predicts the objects should keep their initial placement.}                  
        \label{fig:position_predictions}
\end{figure}

\begin{figure}[H]
    \centering
        \includegraphics{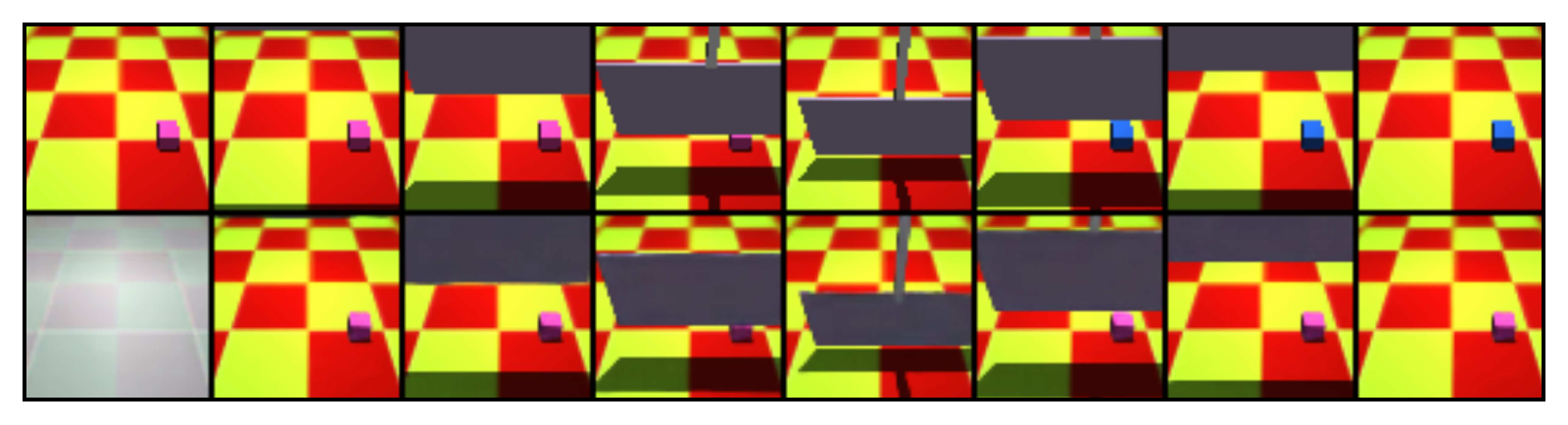}
        \caption{) Network predictions (bottom row) for Unchangeableness Color inconsistent probe (top row).  Even though the cube has changed color after the occluder is lifted, the network predicts it should keep the original color.}                  
        \label{fig:color_predictions}
\end{figure}

\begin{figure}[H]
        \centering
        \includegraphics{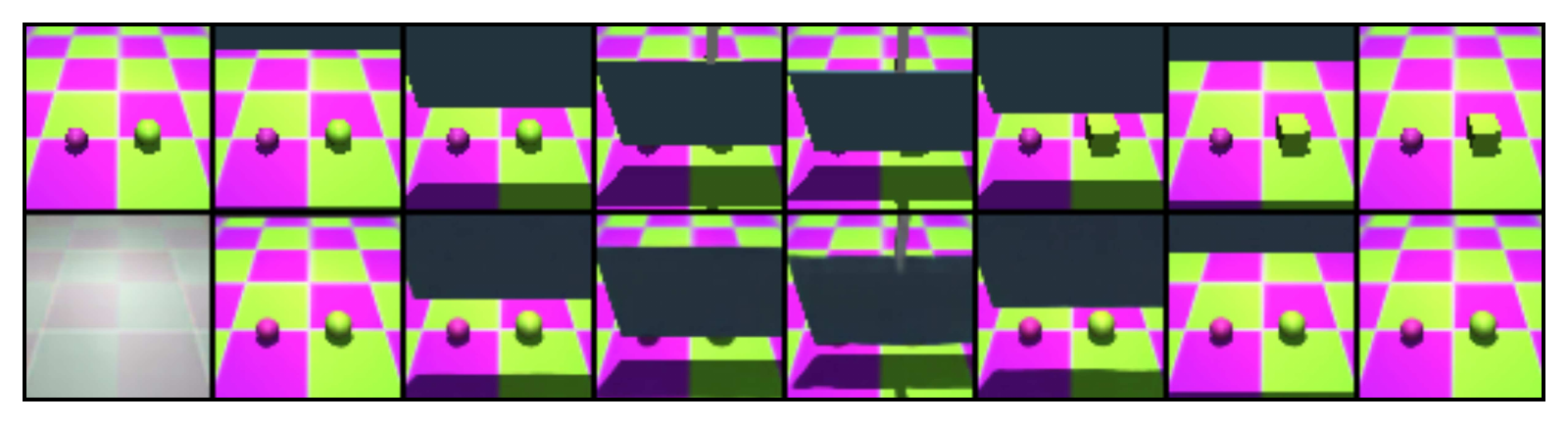}
        \caption{) Network predictions (bottom row) for Unchangeableness Shape inconsistent probe (top row).  Even though the yellow sphere has been changed to a yellow cube after the occluder is lifted, the network predicts it should remain a sphere.}                  
        \label{fig:shape_predictions}
\end{figure}
    
\begin{figure}[H]
        \centering
        \includegraphics{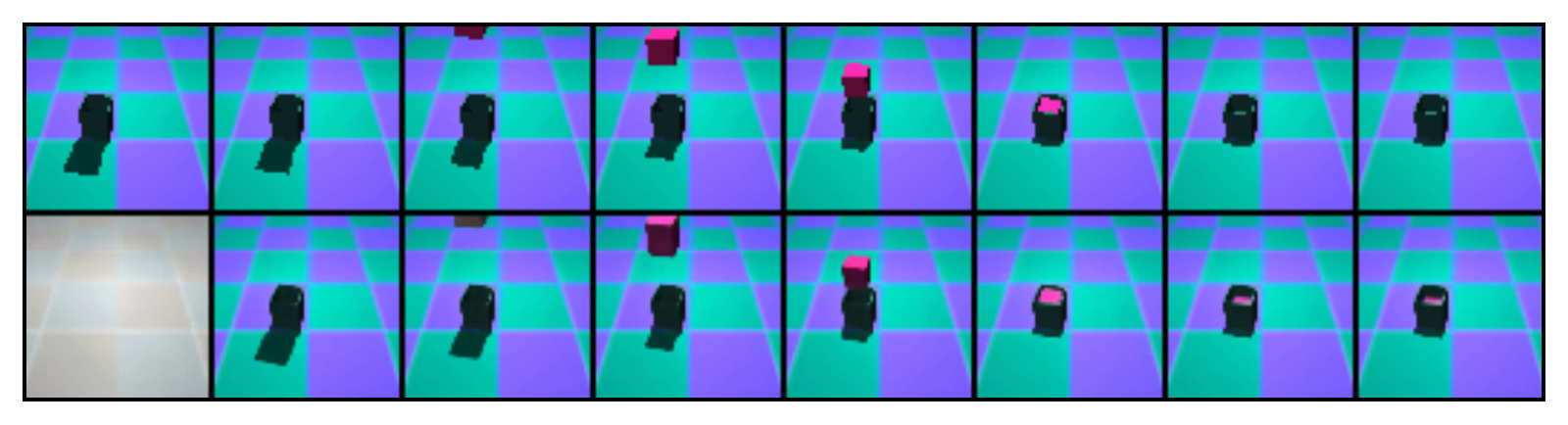}
        \caption{) Network predictions (bottom row) for Object Solidity inconsistent probe (top row).  Even though the cube falls through the container and floor in the inconsistent probe, the model correctly predicts it should remain in the container.}                  
        \label{fig:solidity_predictions}
\end{figure}
    
\begin{figure}[H]
        \centering
        \includegraphics{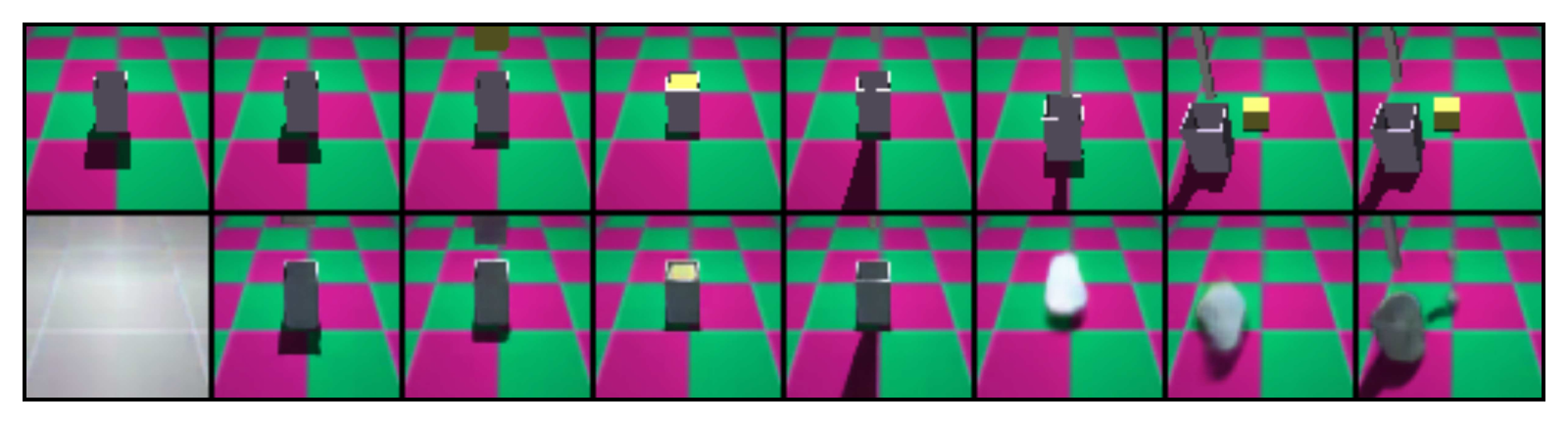}
        \caption{) Network predictions (bottom row) for Containment inconsistent probe (top row).  When the container is moved, the cube that fell into it magically phases through it, remaining in the place the container used to be.  In the first frame after this manipulation, the model correctly predicts nothing should be in the place where the container used to be.  However, the model cannot hold onto this correction for an extended period of time and integrates the inconsistent information into the prior, as seen in the final frame where the space that should be empty has a blurred entity in it.}                  
        \label{fig:containment_predictions}
\end{figure}

\bibliography{punn.bib} \bibliographystyle{icml2017}